\documentstyle[aclap]{article}
\title{\vspace{-0.5in}Recycling Lingware in a Multilingual MT System}

\author{\ \\
        Manny Rayner\\
        David Carter\\ 
        \ \\
        \ \\
        \ \\
        SRI International\\
        Suite 23, Millers Yard\\
        Cambridge CB2 1RQ\\
        United Kingdom\\
        \ \\
        {\tt manny@cam.sri.com}\\
        {\tt dmc@cam.sri.com}\\
        \And
        \ \\
        Ivan Bretan\\
        Robert Eklund\\
        Mats Wir\'en\\
        \ \\
        \ \\
        Telia Research AB\\
        Spoken Language Processing\\
        S-136\,80 Haninge\\
        Sweden\\
        \ \\
       {\tt Ivan.P.Bretan@telia.se}\\
       {\tt Robert.H.Eklund@telia.se}\\
       {\tt Mats.G.Wiren@telia.se}\\
       \And
        \rm Steffen Leo Hansen\\
        Sabine Kirchmeier-Andersen\\
        Christina Philp\\
        Finn S\o rensen\\
        Hanne Erdman Thomsen\\
        \ \\
        Handelsh\o jskolen i K\o benhavn\\
	Institut for Datalingvistik\\
	Dalgas Have 15\\
	DK-2000 Frederiksberg\\
        Denmark\\
        \ \\
	{\tt slh.id@cbs.dk}}

\begin{document}
\bibliographystyle{fullname}
\maketitle

\vspace{-0.5in}

\begin{abstract}

We describe two methods relevant to multi-lingual machine
translation systems, which can be used to port linguistic data
(grammars, lexicons and transfer rules) between systems used for
processing related languages. The methods are fully implemented 
within the Spoken Language Translator system, and were used to
create versions of the system for two new language pairs using
only a month of expert effort.

\end{abstract}

\section{Introduction}

The basic idea of this paper is simple and uncontroversial. All natural
languages are in some sense similar (some are obviously very similar),
so software written to process one language ought to some
extent to be applicable to other languages. If the languages $L_1$ and
$L_2$ are similar enough, then it should be easier to recycle software
applicable to $L_1$ than to rewrite it from scratch for $L_2$.

This paper describes two related approaches in this general direction,
which have been successfully applied within the Spoken Language
Translator (SLT) project \cite{RaynerCarter:97}. The first is the most
obvious: we start with a functioning grammar and lexicon for $L_1$,
and port it to the similar language $L_2$. This is not, of course, a
novel idea, but we think that we have refined it in a number of
ways. In particular, we show that it is practically feasible in the
case of sufficiently close languages to generalize an existing grammar
for $L_1$ to cover both $L_1$ {\em and} $L_2$ (i.e.\ produce a single
grammar which through the setting of a single parameter becomes valid
for either language). We also describe a method which makes it
possible to port the language-dependent lexicon for $L_1$ so as to
maximize sharing of data between the systems for the two languages.

The second idea is specifically related to translation. Suppose we
have already developed sets of transfer rules for the two
language-pairs $L_1 \rightarrow L_2$ and $L_2 \rightarrow L_3$.
We describe a method which allows us to compose the two sets of
rules off-line to create a new set for the pair $L_1 \rightarrow L_3$.

Both methods might be said to operate according to the principle
memorably described by Mary Poppins as ``Well begun is half done''.
They do not solve either problem completely, but automatically take
care of most of the drudgery before any human has to become
involved. In each case, the initial result is a machine-written set of
linguistic data (lexicon entries and transfer rules) which is not
quite adequate as it stands; a system expert can however clean it up
into satisfactory shape in a small fraction of the time that would
have been required to write the relevant rules and lexicon entries
from scratch.

The practical experiments we describe have been carried out using
versions of the SLT system involving the languages English, French,
Swedish and Danish. Initial results are extremely promising. In
particular, we were able to combine both methods to create fairly
credible Swedish-to-French and English-to-Danish spoken language
translation systems\footnote{In fact we do not currently use a Danish
speech synthesizer, but it would be straightforward to incorporate
one.}  
using only a few person-weeks of expert effort.

The rest of the paper is structured as
follows. Section~\ref{Section:SLTOverview} gives a very brief overview
of the relevant aspects of the SLT system.
Section~\ref{Section:CloseLangPort} describes the methods we have
developed for porting linguistic descriptions between closely related
languages. Section~\ref{Section:TransferComposition} summarizes the
transfer composition method. Section~\ref{Section:Experiments}
describes preliminary experiments. 

\section{An overview of the SLT system}
\label{Section:SLTOverview}

The SLT system has been described in detail elsewhere (most recently
\cite{RaynerBouillon:95,RaynerCarter:97}), so this section will only
provide the minimum information necessary to understand the main body
of the paper.

The language-processing (translation) part of the system is supplied
with N-best sentence hypotheses by the system's recognizer, and itself
uses a hybrid architecture, which combines rules and trainable
statistical models. To summarize the argument from
\cite{RaynerBouillon:95}, there are good reasons for requiring both
these components to be present. Rules are useful for capturing many
kinds of regular linguistic facts that are independent of any
particular domain of application, prime examples being grammatical
agreement and question-formation. In contrast, there are other types
of phenomena which intuitively are more naturally conceptualized as
idiosyncratic and domain-dependent: the most obvious examples here are
word-choice problems.

The system uses two translation mechanisms, applied bottom-up in
parallel \cite{RaynerCarter:97}. The primary, rule-based translation
mechanism performs transfer at the level of Quasi-Logical Form (QLF),
a type of predicate/argument style notation \cite{AlshawiEA:91}. The
source- and target-language grammars provide a declarative definition
of a many-to-many mapping between surface form and QLF. The grammars
are domain-independent, and can be compiled to run efficiently either
in the direction surface form $\rightarrow$ QLF (analysis) or QLF
$\rightarrow$ surface form (generation). In transfer,
unification-based rules are used to define a space of possible
candidate translations; domain-dependent, statistically trained
preferences then select the most preferred candidate translation. This
division of effort has the important consequence of allowing the
transfer rules to be fairly simple, since much of the complexity is
``factored out'' into the trained preferences.

In order to deal with the brittleness inherent in any rule-based
system, a second, much less sophisticated translation method is also
used, which simply maps surface phrases from the source language into
possible target-language counterparts. We refer to the backup method
as ``word-to-word'' (WW) translation. The two methods are combined,
roughly speaking, by using rule-based QLF transfer to translate as
much as possible, filling in any gaps with applications of the WW
rules.

The parts of the system of central interest here are the rule-based
components, in particular the morphologies, grammars, lexica and
transfer rules.  Morphologies are written in a variant of two-level
morphology \cite{Carter:95}, and grammars in a unification-based
formalism \cite{CLE}.  The lexicon for each language is divided into
three main parts:
\begin{itemize}
\item Domain-independent function (closed class) word entries are
written directly in terms of definitions of suitable feature-value
assignments, and can from a software-engineering standpoint be
regarded as part of the grammar.

\item A collection of language-dependent but domain-independent macros
define the feature-value assignments needed for each type of regular
content-word, e.g.\ ``count noun'', ``transitive verb'' and so on.
These macros are called {\it paradigm macros}.

\item Content word entries, which in general may be domain-dependent,
are defined in terms of these lexical macros.  An entry of this kind
contains the following information: the name of the relevant macro,
the base surface form of the word, the associated logical-form (QLF)
constant, and if necessary a pointer to the correct inflectional type
(conjugation or declension).
\end{itemize}
Structurally, transfer rules have much in common with lexicon
entries. (Bear in mind that conventional bilingual and monolingual
dictionaries have similar structures too). A small set of
domain-independent transfer rules encode cross-linguistic divergences
significant enough that they need to be represented at the QLF level:
these rules may contain arbitrary pieces of QLF form. The majority of
the transfer rules, however, are ``atomic-atomic'': they associate a
logical-form constant from the source language with one or more
logical-form constants from the target language. Transfer rules of
this type have a close connection with the macro-defined monolingual
content-word lexicon, and may also be domain-dependent.

\section{Porting grammars and lexica between closely related languages}
\label{Section:CloseLangPort} 

The original version of the Core Language Engine had a single language
description for English, written by hand from scratch
\cite{CLEEngCoverage,ATISEngCoverage}. Subsequently, language
descriptions have been developed for Swedish \cite{GambackRayner:92},
French and Spanish \cite{RaynerCarterBouillon:96}. In each of these
cases, the new language description was created by manually editing
the relevant files for the closest existing language. (The Swedish and
French systems are modified versions of the original English one; the
Spanish system is modified from the French one). There are however
some serious drawbacks to this approach. Firstly, it requires a
considerable quantity of expert effort; secondly, there is no
mechanism for keeping the resulting grammars in step with each
other. Changes are often made to one grammar and not percolated to the
other ones until concrete problems show up in test suites or demos.
The net result is that the various grammars tend to drift steadily further
apart.

When we recently decided to create a language description for Danish,
we thought it would be interesting to experiment with a more
principled methodology, which explicitly attempts to address the problems
mentioned above. The conditions appeared ideal: we were porting from
Swedish, Swedish and Danish being an extremely closely related
language pair. The basic principles we have attempted to observe are the
following:
\begin{itemize}
\item Whenever feasible, we have tried to arrange things so that the
linguistic descriptions for the two languages consist of shared files.
In particular, the grammar rules files for the two languages are
shared. When required, rules or parts of rules specific to one
language are placed inside macros whose expansion depends on the
identity of the current language, so that the rule expands when loaded
to an appropriate language-specific version.

\item When files cannot easily be shared (in particular, for the
content-word lexica), we define the file for the new language in terms
of declarations listing the explicit differences against the
corresponding file for the old language. We have attempted to make the
structure of these declarations as simple as possible, so that they
can be written by linguists who lack prior familiarity with the system
and its notation.
\end{itemize}
Although we are uncertain how much generality to claim for the results
(Swedish and Danish, as already noted, are exceptionally close), we
found them encouraging. Four of the 175 existing Swedish grammar
rules turned out to be inapplicable to Danish, and two had to be 
replaced by corresponding Danish rules. Five more rules had to be
parameterized by language-specific macros.
Some of the morphology rules needed to be rewritten, but this only
required about two days of effort from a system specialist working
together with a Danish linguist. The most significant piece of work,
which we will now describe in more detail, concerned the lexicon.

Our original intuition here was that the function-word lexicon and the
paradigm macros (cf Section~\ref{Section:SLTOverview}) would be
essentially the same between the two languages, except that the
surface forms of function words would vary. To put it slightly
differently, we anticipated that it would make sense as a first
approximation to say that there was a one-to-one correspondence
between Swedish and Danish function-words, and that their QLF
representations could be left identical. This assumption does indeed
appear to be borne out by the facts. The only complication we have
come across so far concerns definite determiners: the feature-value
assignments between the two languages need to differ slightly in order
to handle the different rules in Swedish and Danish for
determiner/noun agreement. This was handled, as with the grammar
rules, by introduction of a suitable call to a language-specific
macro.

With regard to content words, the situation is somewhat
different. Since word choice in translation is frequently determined
both by collocational and by semantic considerations, it does not make
as much sense to insist on one-to-one correspondences and identical
semantic representations.  We consequently decided that content-words
would have a language-dependent QLF representation, so as to make it
possible to use our normal strategy of letting the Swedish-to-Danish
translation rules in general be many-to-many, with collocational
preferences filtering the space of possible transfers.

The remarks above motivate the concrete lexicon-porting strategy which
we now sketch. All work was carried out by Danish linguists who had a
good knowledge of computational linguistics and Swedish, but
no previous exposure to the system.  The starting point was to write a
set of word-to-word translation rules (cf
Section~\ref{Section:SLTOverview}), which for each Swedish surface
lexical item defined a set of possible Danish translations. The
left-hand side of each WW rule specified a Swedish surface word-form
and an associated grammatical category (verb, noun, etc), and the
right-hand side a possible Danish translation. An initial ``blank''
version of the rules was created automatically by machine analysis of
a corpus; the left-hand side of the rule was filled in correctly, and
a set of examples taken from the corpus was listed above. The linguist
only needed to fill in the right-hand side appropriately with
reference to the examples supplied.

The next step was to use the word-to-word rules to induce a Danish
lexicon. As a first approximation, we assumed that the possible
grammatical (syntactic/semantic) categories of the word on the
right-hand side of a WW rule would be the same as those of the word on
its left-hand side. (Note that in general a word will have more than
one lexical entry). Thus lexicon entries could be copied across from
Swedish to Danish with appropriate modifications. In the case of
function-words, the entry is copied across with only the surface form
changed. For content-words, the porting routines query the linguist
for the additional information needed to transform each specific item
as follows.

If the left-hand (Swedish) word belongs to a lexical category subject
to morphological inflection, the linguist is asked for the root form
of the right-hand (Danish) word and its inflectional pattern. If the
inflectional pattern is marked as wholly or partly irregular (e.g.\
with strong verbs), the linguist is also queried for the values of
the relevant irregular inflections. All requests for lexical
information are output in a single file at the end of the run,
formatted for easy editing. This makes it possible for the linguist to
process large numbers of information requests quickly and efficiently,
and feed the revised declarations back into the porting process in an
iterative fashion.

One particularly attractive aspect of the scheme is that transfer
rules are automatically generated as a byproduct of the porting
process. Grammar rules and function-words are regarded as 
interlingual; thus for each QLF constant $C$ involved in the
definition of a grammar rule or a function-word definition, the system
adds a transfer rule which maps $C$ into itself. Content-words are not
interlingual. However, since each target lexical entry $L$ is created
from a source counterpart $L^\prime$, it is trivial to create
simultaneously a transfer rule which maps the source QLF
constant associated with $L^\prime$ into the target QLF constant
associated with $L$.

\section{Transfer composition}
\label{Section:TransferComposition}

The previous sections have hopefully conveyed some of the flavour of
our translation framework, which conceptually can be thought of as
half-way between transfer and interlingua. We would if possible like
to move closer to the interlingual end; however, the problems touched
on above mean that we do not see this as being a realistic short-term
possibility. Meanwhile, we are stuck with the problem that dogs all
multilingual transfer-based systems: the number of sets of transfer
rules required increases quadratically in the number of system
languages.  Even three languages are enough to make the problem
non-trivial.

In a recent paper \cite{RaynerEtAl:96}, we described a novel approach
to the problem which we have implemented within the SLT system.
Exploiting the declarative nature of our transfer formalism, we
compose (off-line) existing sets of rules for the language pairs $L_1
\rightarrow L_2$ and $L_2 \rightarrow L_3$, to create a new set of
rules for $L_1 \rightarrow L_3$. It is clear that this can be done for
rules which map atomic constants into atomic constants.  What is less
obvious is that complex rules, recursively defined in terms of
translation of their sub-constituents, can also be composed. The
method used is based on program-transformation ideas taken from logic
programming, and is described in detail in the earlier paper. Simple
methods, described in the same paper, can also be used to compose an
approximate transfer preference model for the new language-pair.

The rule composition algorithm is not complete; we strongly suspect
that, because of recursion effects, the problem of finding a complete
set of composed transfer rules is undecidable. But in practice, the
set of composed rules produced is good enough that it can be improved
quickly to an acceptable level of performance. Our methodology for
performing this task makes use of rationally constructed, balanced
domain corpora to focus the effort on frequently occurring problems
\cite{RaynerBouillonCarter:95}. It involves making declarations to
reduce the overgeneration of composed rules; adding hand-coded rules
to fill coverage holes; and adjusting preferences. The details
are reported in \cite{RaynerEtAl:96}.

\section{Experiments}
\label{Section:Experiments}

We will now present results for concrete experiments, where we applied
the methods described above so as to rapidly construct translation
systems for two new language pairs.  All of the translation modules
involved operate within the same Air Travel Inquiry (ATIS;
\cite{ATIS}) domain as other versions of SLT, using a vocabulary of
about 1\,500 source-language stem entries, and have been integrated
into the main SLT system to produce versions which can perform
credible translation of spoken Swedish into French and spoken English
into Danish respectively.

\subsection{Swedish $\rightarrow$ English $\rightarrow$ French}
\label{Section:SweEngFre}

This section describes an exercise which involved using transfer
composition to construct a Swedish $\rightarrow$ French translation
system by composing Swedish $\rightarrow$ English and English
$\rightarrow$ French versions of the system. The total expert effort
was about two person-weeks.  We start by summarizing results, and then
sketch the main points of the manual work needed to adjust the
composed rule-sets.

We used a corpus of 442 previously unseen spoken utterances, and
processed the N-best lists output for them by the speech recognizer.
The results are as given in Table \ref{swefretab}; for comparison,
we also give the results for English $\rightarrow$ Swedish, the
language pair to which we have devoted the most effort (and which
does not involve any transfer composition).

\begin{table}
\begin{center}
\begin{tabular}{|l|c|c|} \hline \hline
             &  Swe $\rightarrow$ Fre  & Eng $\rightarrow$ Swe \\ \hline \hline
Fully acceptable       & 29.4\% & 56.5\%  \\ \hline
Unnatural style        & 16.3\% &  7.75\% \\ \hline
Minor syntactic errors & 15.2\% & 11.75\% \\ \hline \hline
Major syntactic errors & 2.0\%  &  4.75\% \\ \hline
Partial translation    & 7.0\%  &  8.75\% \\ \hline \hline
Nonsense               & 22.9\% &  5.0\%  \\ \hline
Bad translation        & 7.0\%  &  4.0\%  \\ \hline
No translation         & 0.2\%  &  1.5\%  \\ \hline \hline
\end{tabular}
\end{center}
\caption{Translation results for Swedish $\rightarrow$ French and 
English $\rightarrow$ Swedish on unseen speech data}
\label{swefretab}
\end{table}

Thus almost 30\% (top row) of the translations produced were
completely acceptable, with another 30\% or so (rows 2-3) having only
minor problems, giving a total of 60\% that would probably be
acceptable in practical use. A further 9\% (rows 4-5) contained major
errors but also some correct information, while nearly all the
remaining 30\% (bottom 3 rows) were clearly unacceptable, consisting
either of nonsense or of a translation that made some sense but was
wrong. The reasons for these 30\% of outright failures, compared to
only about 10\% for English $\rightarrow$ Swedish, are firstly, that
recognizer performance is slightly less good for Swedish than for
English, owing to less training data being available; second, that
Swedish and French differ more than English and Swedish do; thirdly,
that transfer rules for both the component pairs (Swedish
$\rightarrow$ English and English $\rightarrow$ French) have had much
less work devoted to them than English $\rightarrow$ Swedish; and last
but not least, of course, that transfer composition is being used.

When cleaning up the automatically composed Swedish $\rightarrow$
French rule-set, the task on which we spent most effort was that of
limiting overgeneration of composed transfer rules.  The second most
important task was manual improvement of the composed transfer
preference model. The methods used are described in more detail in
\cite{RaynerEtAl:96}.

\subsection{English $\rightarrow$ Swedish $\rightarrow$ Danish}

This section briefly describes a second series of experiments, in
which we converted an English $\rightarrow$ Swedish system into an
English $\rightarrow$ Danish system using the methods described
earlier. The total investment of system expert effort was
again around two person-weeks.

About half the effort was used to port the Swedish language
description to Danish, employing the methods of
Section~\ref{Section:CloseLangPort}. After this, we carried out two
rounds of testing and bug-fixing on the Swedish $\rightarrow$ Danish
translation task. For this, we used a Swedish representative
corpus, containing 331 sentences representing 9\,385 words from the
original Swedish corpus. These tests uncovered a number of new
problems resulting from previously unnoted divergences between the
Swedish and Danish grammars. About half the problems disappeared after
the addition of 20 or so small hand-coded adjustments to the
morphology, function-word lexicon, transfer rules and transfer
preferences.

After the second round of bug-fixing, 95\% of the Swedish sentences
received a Danish translation, and 79\% a fully acceptable
translation.  (When measuring results on representative corpora, we
count coverage in terms of ``weighted scores''. The weight assigned to
sentence is proportional to the number of words it represents in the
original corpus: that is, its length in words times the number of
sentences it represents).  Most of the translation errors that did
occur were minor ones. Finally, we composed the English $\rightarrow$
Swedish and Swedish $\rightarrow$ Danish rules to create a English
$\rightarrow$ Danish rule-set, and used this, after a day's editing by
an expert, to test English $\rightarrow$ Danish translation using a
representative text corpus (we will present results for unseen speech
input at the workshop).  Our results, using the same scheme as above,
were as given in Table \ref{engdantab}.

\begin{table}
\begin{center}
\begin{tabular}{|l|c|} \hline \hline
             &  Eng $\rightarrow$ Dan  \\ \hline \hline
Fully acceptable       & 52.5\% \\ \hline
Unnatural style        & 0.4\% \\ \hline
Minor syntactic errors & 24.4\% \\ \hline \hline
Major syntactic errors & 0.7\% \\ \hline
Partial translation    & 0.0\% \\ \hline \hline
Nonsense               & 0.9\% \\ \hline
Bad translation        & 10.7\% \\ \hline
No translation         & 10.3\% \\ \hline \hline
\end{tabular}
\end{center}
\caption{Translation results for English $\rightarrow$ Danish on
representative text data}
\label{engdantab}
\end{table}

\section{Conclusions and further directions}
\label{Section:Conclusions}

We have demonstrated that it is practically feasible in the case of
sufficiently close languages to generalize an existing grammar for one
language to produce a grammar which, through the setting of a single
parameter, becomes valid for either language. As well as providing
major efficiency gains over writing a grammar for the second language
from scratch, this technique means that subsequent enhancements to the
grammar, in those areas where the characteristics of the two languages
are equivalent, will apply automatically to both of them.

We have also described an algorithm for composition of transfer
rules. We have demonstrated that it can be used to automatically
compose non-trivial sets of transfer rules containing on the order of
thousands of rules, and shown that by small adjustments the
performance can be improved to a level only slightly inferior to that
of a corresponding set of hand-coded rules. Our experience is that the
amount of work involved in using these methods is only a fraction of
that needed to develop similar rules from scratch.

\section*{Acknowledgements}

The Danish-related work reported here was funded by SRI International
and Handelsh\o jskolen i K\o benhavn. Other work was funded by Telia
Research AB under the SLT-2 project. We would like to thank David
Milward and Steve Pulman for helpful comments, and Thierry Reynier for
judging the Swedish-to-French translations.

\end{document}